\documentclass[11pt,a4paper]{article}
\usepackage[hyperref]{acl2019}
\usepackage{times}
\usepackage{latexsym}

\usepackage{url}

\usepackage{booktabs}

\usepackage{bm}
\usepackage{amsmath}
\usepackage{graphicx}
\usepackage{amssymb}
\usepackage{algorithmicx,algorithm}
\usepackage[noend]{algpseudocode}
\usepackage{multirow}

\aclfinalcopy 
\def\aclpaperid{2788} 

\title{Retrieving Sequential Information for Non-Autoregressive \\Neural Machine Translation}

\author{Chenze Shao$^1$$^2$$^3$, Yang Feng$^1$$^2$$^\star$, Jinchao Zhang$^3$, Fandong Meng$^3$, Xilin Chen$^1$$^2$ and Jie Zhou$^3$ \\
$^{1}$ University of Chinese Academy of Sciences\\
$^{2}$ Key Laboratory of Intelligent Information Processing\\
Institute of Computing Technology, Chinese Academy of Sciences (ICT/CAS)\\
$^{3}$ Pattern Recognition Center, WeChat AI, Tencent Inc, China\\
{\tt \{\href{mailto:shaochenze18z@ict.ac.cn}{shaochenze18z}, \href{mailto:fengyang@ict.ac.cn}{fengyang}, \href{mailto:xlchen@ict.ac.cn}{xlchen}\}@ict.ac.cn }\\ {\tt \{\href{mailto:dayerzhang@tencent.com}{dayerzhang}, \href{mailto:fandongmeng@tencent.com}{fandongmeng}, \href{mailto:withtomzhou@tencent.com}{withtomzhou}\}@tencent.com}}
\date{}
\begin{document}
\maketitle
\newcommand\blfootnote[1]{%
\begingroup 
\renewcommand\thefootnote{}\footnote{#1}%
\addtocounter{footnote}{-1}%
\endgroup
}

\begin{abstract}
Non-Autoregressive Transformer (NAT) aims to accelerate the Transformer model through discarding the autoregressive mechanism and generating target words independently, which fails to exploit the target sequential information. Over-translation and under-translation errors often occur for the above reason, especially in the long sentence translation scenario. In this paper, we propose two approaches to retrieve the target sequential information for NAT to enhance its translation ability while preserving the fast-decoding property. 
Firstly, we propose a sequence-level training method based on a novel reinforcement algorithm for NAT (Reinforce-NAT) to reduce the variance and stabilize the training procedure. 
Secondly, we propose an innovative Transformer decoder named FS-decoder to fuse the target sequential information into the top layer of the decoder. 
Experimental results on three translation tasks show that the Reinforce-NAT surpasses the baseline NAT system by a significant margin on BLEU without decelerating the decoding speed and the FS-decoder achieves comparable translation performance to the autoregressive Transformer with considerable speedup. 
\blfootnote{ Joint work with Pattern Recognition Center, WeChat AI, Tencent Inc, China.}
\blfootnote{$\star$ Corresponding Author}

\end{abstract}

\section{Introduction}

Neural machine translation (NMT) models \cite{cho2014learning,sutskever2014sequence,bahdanau2014neural} solve the machine translation problem with the Encoder-Decoder framework and achieve impressive performance on translation quality. Recently, the Transformer model \cite{vaswani2017attention} further enhances the translation performance on multiple language pairs, while suffering from the slow decoding procedure, which restricts its application scenarios. The slow decoding problem of the Transformer model is caused by its autoregressive nature, which means that the target sentence is generated word by word according to the source sentence representations and the target translation history.

\begin{table}[t]
\small
\centering
\begin{tabular}{r|l}
\toprule
\multirow{1}{*}{Src} & und noch tragischer ist , dass es Oxford war $\cdots$\\
\hline

\multirow{1}{*}{Ref} & even more tragic is that it was Oxford $\cdots$\\
\hline
\multirow{1}{*}{NAT} & and more more more more that it was Oxford $\cdots$\\
\hline
\multirow{1}{*}{AR} & and , more tragic , Oxford was $\cdots$\\
\bottomrule
\end{tabular}
\caption{A fragment of a long sentence translation. AR stands for the translation of the autoregressive Transformer. The output of the NAT model contains repeated translations of word `more' and misses the word `tragic'.}
\label{tab:case}
\end{table}

Non-autoregressive Transformer model \cite{gu2017non} is proposed to accelerate the decoding process, which can simultaneously generate target words by discarding the autoregressive mechanism. 
Since the generation of target words is independent, NAT models utilize alternative information such as encoder inputs \cite{gu2017non}, translation results from other systems \cite{lee2018deterministic,guo2018non} and latent variables \cite{kaiser2018fast} as decoder inputs.
Without considering the target translation history, NAT models are weak to exploit the target words collocation knowledge and tend to generate repeated target words at adjacent time steps \cite{wang2019non}. Over-translation and under-translation problems are aggravated and often occur due to the above reasons.
Table \ref{tab:case} shows an inferior translation example generated by a NAT model.
Compared to the autoregressive Transformer, NAT models achieve significant speedup while suffering from a large gap in translation quality due to the lack of target sequential information. 

In this paper, we present two approaches to retrieve the target sequential information for NAT models to enhance their translation ability and meanwhile preserve the fast-decoding property. 
Firstly, we propose a sequence-level training method based on a novel reinforcement algorithm for NAT (\textit{Reinforce-NAT}) to reduce the variance and stabilize the training procedure.
We leverage the sequence-level objectives (e.g., BLEU \cite{papineni2002bleu}, GLEU \cite{wu2017adversarial}, TER \cite{snover2006study}) instead of the cross-entropy objective to encourage NAT model to generate high quality sentences rather than the correct token for each position. 
Secondly, we propose an innovative Transformer decoder named \textit{FS-decoder} to fuse the target sequential information into the top layer of the decoder.
The bottom layers of the FS-decoder run in parallel to keep the decoding speed and the top layer of the FS-decoder can exploit target sequential information to guide the target words generation procedure.

We conduct experiments on three machine translation tasks (IWSLT16 En$\rightarrow$De, WMT14 En$\leftrightarrow$De, WMT16 En$\rightarrow$Ro) to validate our proposed approaches. 
Experimental results show that the Reinforce-NAT surpasses the baseline NAT system by a significant margin on the translation quality without decelerating the decoding speed, and the FS-decoder achieves comparable translation capacity to the autoregressive Transformer with considerable speedup.

\section{Background}
\subsection{Autoregressive Neural Machine Translation}

Given a source sentence $\bm{X}=\{x_1, ..., x_{n}\}$ and a target sentence $\bm{Y}=\{y_1, ..., y{_T}\}$, autoregressive NMT models the translation probability from $\bm{X}$ to $\bm{Y}$ as:
\begin{equation}
\label{eq:auto_prob}
P(\bm{Y}|\bm{X},\theta) = \prod_{t=1}^{T}p(y_t|\bm{y_{<t}},\bm{X},\theta),
\end{equation}
where $\theta$ is a set of model parameters and $\bm{y_{<t}}=\{y_1,\cdots,y_{t-1}\}$ is the translation history. 
Given the training set $D = \{\rm{\mathbf{X}}^{M},\mathbf{Y}^M\}$ with $M$ sentence pairs, the training objective is to maximize the log-likelihood of the training data as:
\begin{equation}
\begin{aligned}
\label{eq:auto_mle}
&\bm{\theta} = \arg\max_{\theta}\{\mathcal{L}(\theta)\}\\
\mathcal{L}(\theta) = \sum_{m=1}^{M}&\sum_{t=1}^{T}\log(p( y _t^m|\bm{ y_}{<t}^m,\bm{X}^m,\theta)),
\end{aligned}
\end{equation}
where the superscript $m$ indicates the m-th sentence in the dataset. During training, golden target words are fed into the decoder as the translation history. During inference, the partial translation generated by decoding algorithms such as greedy search and beam search is fed into the decoder to guide the generation of the next word.

The prominent feature of the autoregressive model is that it requires the target side historical information in the decoding procedure. Therefore target words are generated in the one-by-one style. Due to the autoregressive property, the decoding speed is limited, which 
restricts the application of the autoregressive model.

\subsection{Sequence-Level Training for Autoregressive NMT}
Reinforcement learning techniques \cite{sutton2000policy,Ng1999PolicyIU,sutton1984temporal} have been widely applied to improve the performance of the autoregressive NMT with sequence-level objectives \cite{shen2016minimum,ranzato2015sequence,bahdanau2016actor}. As sequence-level objectives are usually non-differentiable, the loss function is defined as the negative expected reward:
\begin{equation}
\mathrm{L}_\theta=- \sum_{\bm{\mathrm{Y}}=y_{1:T}}{p(\bm{\mathrm{Y}}|\bm{\mathrm{X}},\theta) \cdot r(\bm{\mathrm{Y}})}, \label{eq:loss}
\end{equation}
where $\bm{\mathrm{Y}}=y_{1:T}$ denotes possible sequences generated by the model, and $r(\bm{\mathrm{Y}})$ is the corresponding reward such as BLEU, GLEU and TER for generating sequence $\bm{\mathrm{Y}}$. Enumerating all the possible target sequences is impossible due to the exponential search space, and REINFORCE \cite{williams1992simple} gives an elegant way to estimate the gradient for Eq.(\ref{eq:loss}) via sampling a sequence $\bm{\mathrm{Y}}$ from the probability distribution and estimate the gradient with the gradient of log-probability weighted by the reward $r(\bm{\mathrm{Y}})$:
\begin{equation}
\begin{aligned}
\label{eq:auto_rf}
&\nabla_{\theta}\mathrm{L}_\theta=\\
&- \mathop{\mathbb{E}}\limits_{\bm{\mathrm{Y}}} [\sum_{t=1}^{T}\nabla_{\theta} \log(p(y_{t}|\bm{y_{<t}},\bm{\mathrm{X}},\theta)) \cdot r(\bm{\mathrm{Y}})].
\end{aligned}
\end{equation}

Current reinforcement learning (RL) methods are designed for autoregressive models. Moreover, previous investigations \cite{wu2018study,article} show that the RL-based training procedure is unstable due to its high variance of gradient estimation. 

\subsection{Non-Autoregressive Neural Machine Translation}
Non-autoregressive neural machine translation \cite{gu2017non} is proposed to accelerate the decoding process,  which can simultaneously
generate target words by discarding the autoregressive mechanism. 

The translation probability from $\bm{X}$ to $\bm{Y}$ is modeled as follows:

\begin{equation}
\label{eq:nonauto_prob}
P(\bm{Y}|\bm{X},\theta) = \prod_{t=1}^{T}p(y_t|\bm{X},\theta).
\end{equation}
Given the training set $D = \{\rm{\mathrm{X}}^\mathrm{M},Y^{M}\}$ with $M$ sentence pairs, the training objective is to maximize the log-likelihood of the training data as:
\begin{equation}
\begin{aligned}
\label{eq:nonauto_mle}
&\bm{\theta} = \arg\max_{\theta}\{\mathcal{L}(\theta)\}\\
\mathcal{L}(\theta) = &\sum_{m=1}^{M}\sum_{t=1}^{T}\log(p( y_t^m|\bm{X}^m,\theta)).
\end{aligned}
\end{equation}

During decoding, the translation with maximum likelihood can be easily obtained by taking the word with the maximum likelihood in every time step:
\begin{equation}
\label{eq:argmax}
\hat{y_t} = \arg\max_{y_t}p(y_t|\bm{X},\theta)
\end{equation}

NAT models do not utilize the target translation history, which results in its weakness in exploiting the target words collocation knowledge for generating correct target word sequence under the cross-entropy objective function. 
Compared to autoregressive models, NAT models achieve significant speedup while suffering from a large gap in the translation quality due to the lack of target sequential information.

\section{Approaches}
To retrieve the sequential information for NAT models for enhancing their translation ability and meanwhile preserving the fast-decoding property, we present two approaches: sequence-level training with a reinforcement algorithm for NAT models (Reinforce-NAT) to exploit the sequential information, and a novel Transformer decoder named FS-decoder to fuse sequential information into the top layer.
\subsection{Sequence-Level Training for NAT Models}
Word-level objective functions, such as the cross-entropy loss, focus on generating the correct token in each position, which will be inferior for NATs without the target sequential information.
We propose to encourage NAT models to generate high-quality sentences rather that correct words with the sequence-level training algorithm (Reinforce-NAT). 
\subsubsection*{Algorithm Derivation}
In this section, we present the derivation of Reinforce-NAT and show its low variance and efficiency.
We first introduce the REINFORCE algorithm \cite{williams1992simple} for NAT models. 

In NAT models, with the non-autoregressive translation probability defined in Eq.(\ref{eq:nonauto_prob}), the gradient of the expected loss is:
\begin{equation}
\begin{aligned}
\label{eq:nonauto_grad}
\nabla_{\theta}\mathrm{L}_\theta=-\sum_{\bm{\mathrm{Y}}}\nabla_{\theta}\prod_{t=1}^{T}p(y_t|\bm{X},\theta) \cdot r(\bm{\mathrm{Y}}).
\end{aligned}
\end{equation}

Directly applying the REINFORCE algorithm to Eq.(\ref{eq:nonauto_grad}) will make the gradient update in every postion guided by the same sentence reward $r(\bm{\mathrm{Y}})$, which is similar to the method for autoregressive models and is unstable during training. Instead, for NAT models, Eq.(\ref{eq:nonauto_grad}) can be further reduced to the following form, which is the gradient of target words probability weighted by their corresponding expected rewards\footnote{The proof is provided in the appendix}:
\begin{equation}
\begin{aligned}
\label{eq:nonauto_reduce}
\nabla_{\theta}\mathrm{L}_\theta=-\sum_{t=1}^{T}\sum_{y_t} \nabla_{\theta} p(y_{t}|\bm{\mathrm{X}},\theta) \cdot r(y_t),
\end{aligned}
\end{equation}
where $r(y_t)$ is the expected reward when $y_t$ is fixed:
\begin{equation}
\begin{aligned}
\label{eq:reward}
&r(y_t) = \mathop{\mathbb{E}}\limits_{y_{1:t-1}}\mathop{\mathbb{E}}\limits_{y_{t+1:T}}r(\bm{\mathrm{Y}}).
\end{aligned}
\end{equation}
In Eq.(\ref{eq:nonauto_reduce}), the predicted word $y_t$ in position $t$ is evluated by its corresponding expected reward $r(y_t)$, which is more accurate than the sentence reward $r(\bm{\mathrm{Y}})$. The $r(y_t)$ can be estimated by Monte Carlo sampling, as illustrated in algorithm \ref{alg:2}. Specifically, we fix $y_t$ in position $t$ and sample other words from the probability distribution $p(\cdot|\bm{\mathrm{X}},\theta))$ for $n$ times. The estimated value of $r(y_t)$ is the average reward of the $n$ sampled sentences. Notice that the expected reward $r(y_t)$ can be estimated without running the decoder for multiple times, which is a major advantage of NAT models in sequence-level training.

The gradient in Eq.(\ref{eq:nonauto_reduce}) can be estimated with REINFORCE \cite{williams1992simple}:
\begin{equation}
\begin{aligned}
\label{eq:nonauto_rf}
\nabla_{\theta}\mathrm{L}_\theta=- \sum_{t=1}^{T}\mathop{\mathbb{E}}_{y_t}[\nabla_{\theta} \log(p(y_{t}|\bm{\mathrm{X}},\theta)) \cdot r(y_t)].
\end{aligned}
\end{equation}

\begin{algorithm}[t]
\caption{Estimation of $r(y_t)$} 
\label{alg:2}
\hspace*{0.02in} {\bf Input:} 
the output probability distribution $p(\cdot|\bm{\mathrm{X}},\theta))$, $t$, $y_t$, $T$, sampling times $n$\\
\hspace*{0.02in} {\bf Output:} 
estimate of $r(y_t)$
\begin{algorithmic}[1]
\State $r$ = $0$, $i$ = $0$
\For{$i < n$}
  \State sample $\tilde{\bm{y}}_{1:t-1}$, $\tilde{\bm{y}}_{t+1:T}$ from $p(\cdot|\bm{\mathrm{X}},\theta))$
  \State $\tilde{\bm{\mathrm{Y}}}$ = $\{\tilde{\bm{y}}_{1:t-1}$, $y_t$, $\tilde{\bm{y}}_{t+1:T}\}$
  \State $r$ += $r(\tilde{\bm{\mathrm{Y}}})$
  \State $i$ += 1
\EndFor
\State $r$ = $r/n$
\State \Return $r$
\end{algorithmic}
\end{algorithm}

Eq.(\ref{eq:nonauto_rf}) corresponds to a gradient estimation method through sampling a target word $y_t$ and the gradient of the log-probability of $y_t$ weighted by reward $r(y_t)$ is utilized to estimate the expected gradient over the vocabulary. Though the estimation is unbiased, the gradient estimator still suffers from high variance. The variance can be eliminated by traversing the whole vocabulary, but it is unaffordable due to the huge vocabulary size.

The probability distribution over the target vocabulary is usually a centered distribution where the top-ranking words occupy the central part of the distribution, and the softmax layer ensures that other words with small probabilities have small gradients\footnote{In the softmax layer, the gradient is proportional to the output probability}. Hence the variance will be effectively reduced if we can eliminate the variance from top-ranking words. This motivates us to compute gradients of the top-ranking words accurately and estimate the rest via the REINFORCE algorithm. 

We can build an unbiased estimation of Eq.(\ref{eq:nonauto_reduce}) by traversing top-$k$ words and estimating the rest via one sampling:
\begin{equation}
\begin{aligned}
\label{eq:rfnat}
&\nabla_{\theta}\mathrm{L}_\theta=-\sum_{t=1}^{T}(\sum_{y_t \in T_K}\nabla_{\theta} p(y_{t}|\bm{\mathrm{X}},\theta)\cdot r(y_t)\\
&+(1-P_k) \cdot \mathop{\mathbb{E}}_{y_t\sim \tilde p}[\nabla_{\theta} \log(p(y_{t}|\bm{\mathrm{X}},\theta)) \cdot r(y_t)]).
\end{aligned}
\end{equation}

Algorithm \ref{alg:1} illustrates the proposed method. Although this algorithm will lead to multiple estimations of the expected reward $r(y_t)$, the training cost is relatively low for the reason that the independent generation of target words makes NAT models efficient in estimating the expected reward, which will be either very expensive \cite{yu2017seqgan} or biased \cite{bahdanau2016actor} for autoregressive models.
\begin{algorithm}[t]
\caption{Reinforce-NAT} 
\label{alg:1}
\hspace*{0.02in} {\bf Input:} 
the output probability distribution $p(\cdot|\bm{\mathrm{X}},\theta))$, traversing count $k$, sample times $n$ \\
\hspace*{0.02in} {\bf Output:} 
estimate of $\nabla_{\theta} \mathrm{L}_\theta$ in position $t$ according to Eq.(\ref{eq:rfnat})
\begin{algorithmic}[1]
\State $T_K$ = \{words ranking top-$k$ in $p(\cdot|\bm{\mathrm{X}},\theta))$\}
\State  $\nabla_{\theta} \mathrm{L}_\theta=0$, $\tilde p = p$, $P_k=0$
\For{$y_t$ in $T_K$}
  \State estimate $r(y_t)$ by algorithm \ref{alg:2} with sample times $n$
  \State $\nabla_{\theta} \mathrm{L}_\theta$ -= $\nabla_{\theta} p(y_{t}|\bm{\mathrm{X}},\theta)\cdot r(y_t)$
  \State $\tilde p(y_{t}|\bm{\mathrm{X}},\theta)$ = 0
  \State $P_k$ += $p(y_{t}|\bm{\mathrm{X}},\theta)$
\EndFor
\State normalize $\tilde p(\cdot|\bm{\mathrm{X}},\theta)$
\State sample $y_t$ from $\tilde p(\cdot|\bm{\mathrm{X}},\theta)$
\State estimate $r(y_t)$ by algorithm \ref{alg:2} with sample times $n$
\State $\nabla_{\theta} \mathrm{L}_\theta$ -= $(1 - P_k) \cdot \nabla_{\theta} \log(p(y_{t}|\bm{\mathrm{X}},\theta))\cdot r(y_t)$
\State \Return $\nabla_{\theta} \mathrm{L}_\theta$
\end{algorithmic}
\end{algorithm}
\subsubsection*{Reinforce-NAT}
To give the clear description, we firstly define symbols in Algorithm \ref{alg:1}:

1) $p(\cdot|\bm{\mathrm{X}},\theta))$ is the output probability distribution generated by the decoder on the target vocabulary at time $t$. 
2) $T_K$ is the set of target words with top-$k$ probabilities.
3) $P_k$ is the sum of probabilities in $T_K$, 
4) $\tilde p$ is the normalized probability distribution after removing probabilities of words in $T_K$.

The algorithm takes the output probability distribution $p$, the traversing count $k$ and the sampling times $n$ as input and output the gradient estimation at step $t$. We divide the gradient estimation procedure at step $t$ into two parts: \emph{traversing} and \emph{sampling}. 

The algorithm firstly builds the set $T_K$ with words ranking top-$k$ in probability (line $1$), then estimates expected rewards for words in $T_K$ by algorithm \ref{alg:2} (line $3$, line $4$).
The accumulated gradient in $T_K$ are obtained by traversing the words in $T_K$ and accumulating gradients of their probability functions, which are weighted by corresponding rewards (line 5).\\
\indent{}After the traversing procedure for accumulating gradients for words in $T_K$, the algorithm estimates the expected gradient for words that are not in $T_K$ in the \emph{sampling} procedure.
The algorithm obtains the probability distribution $\tilde p$ over the rest of words through masking probabilities of words in the $T_k$ (line $6$, line$8$).
A word $y_t$ from the distribution $\tilde p$ (line $9$) is sampled to compute the gradient of the log-probability of $y_t$ and then estimate the reward of $r(y_t)$.
The weight for this estimation is $1-P_k$, where $P_k$ is the sum of probabilities in $T_K$. 
Finally, the estimated gradient is the sum of gradients from Top-k words and the sampled word.  (line $11$).\\
\indent{}In a word, the algorithm aims to traverse gradients of important words since they can dominate the gradient estimation, and estimate the gradient of less important words via one sampling.

\subsection{Fuse Sequential Information}

We propose an innovative Transformer decoder named FS-decoder to fuse the target sequential information into the top layer of the decoder.
The FS-decoder consists of four parts: bottom layers, the fusion layer, the top layer and the softmax layer. In the decoder, we parallelize bottom layers in an non-autoregressive way to accelerate the model but serialize the top layer in an autoregressive way to enhance the translation quality. 
The teacher forcing algorithm \cite{williams1989learning} is applied in the training where target embeddings are directly fed to the fusion layer. 
During decoding, FS-decoder only needs to run the top layer autoregressively.

\begin{figure}[tb]
  \begin{center}
    \includegraphics[scale=0.25]{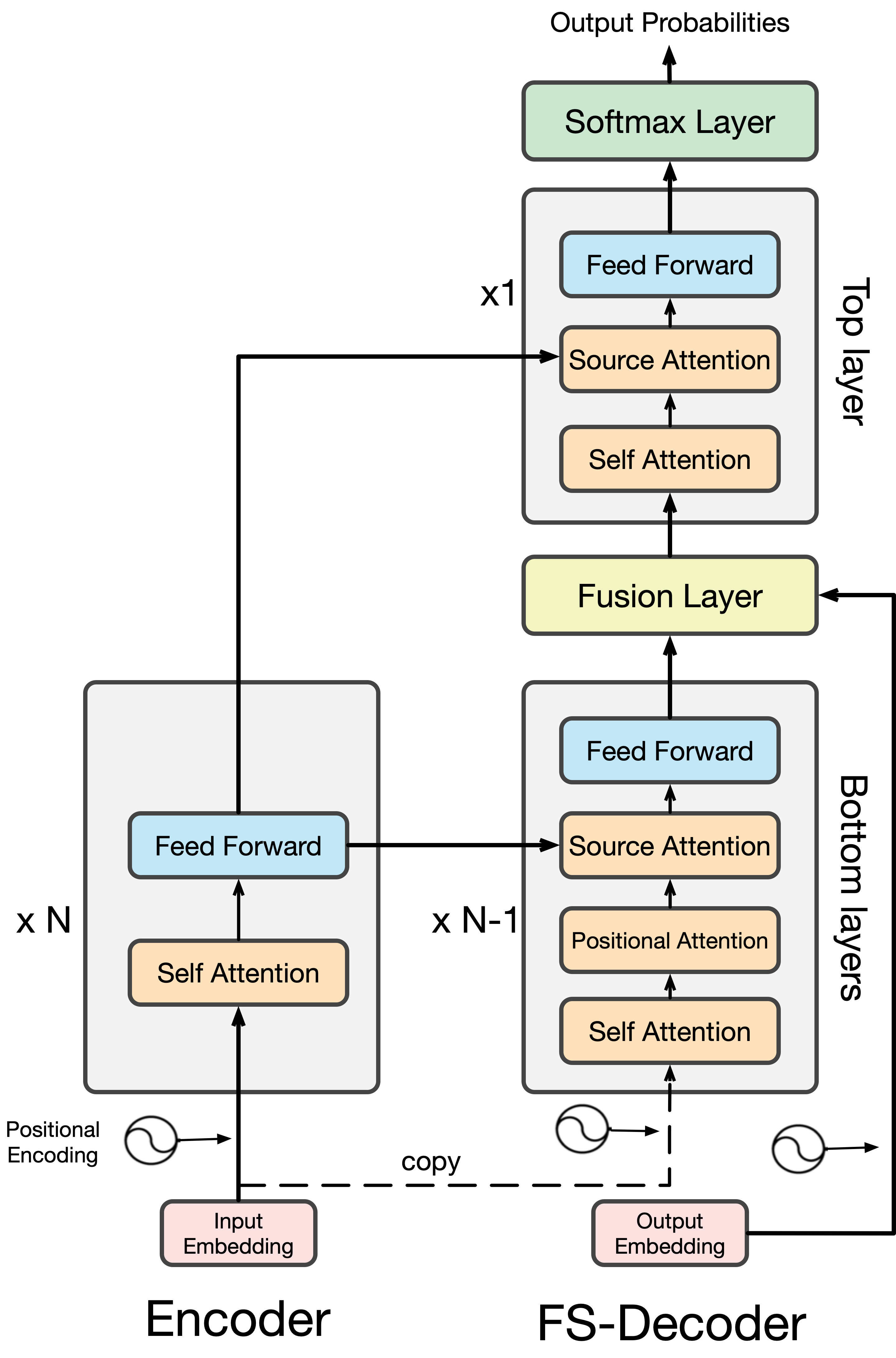}
    \caption{The architecture of FS-decoder. The decoder consists of $n-1$ bottom layers, the fusion layer, the top layer and the softmax layer.}
    \label{fig:2}
  \end{center}
  \vspace{-1em}
\end{figure}

We illustrate the model in figure \ref{fig:2} and describe the detailed architecture of the FS-decoder in the following. Assume that the original Transformer has $n$ decoder layers, the source sentence has length $T_s$, the target sentence has length $T$, and the predicted target length is $T^{'}$. Here we directly look up the source-target length dictionary to predict the target length.

\indent{}\textbf{Bottom Layers.}
The decoder of FS-decoder contains $n$-$1$ bottom layers, which are identical to the decoder layers of NAT models \cite{gu2017non}. Each layer consists of four sub-layers: the self-attention layer, the positional attention layer, the source side attention layer and the position-wise feed-forward layer. The inputs for bottom decoders $\rm{\mathbf{X}}^{'}$ are uniformly copied \cite{gu2017non} from the source input ${\rm{\mathbf{X}}}$ where each decoder input in position $t$ is a copy of the source input in position Round$(T^{'}t/T_s)$:
\begin{equation}
\begin{aligned}
\rm{\mathbf{X}}^{'} = Uniform(\rm{\mathbf{X}}).
\end{aligned}
\end{equation}
The bottom layers take the inputs $\rm{\mathbf{X}}^{'}$ and output the hidden states $\rm{\mathbf{H}}^{'}$ with the same length $T^{'}$.

\indent{}\textbf{Fusion Layer.}
The fusion layer is a linear transformation layer with a ReLU activation, which fuses the outputs from bottom layers $\rm{\mathbf{H}}^{'}$ and target embeddings $\rm{\mathbf{Y}}$ in each position $t$ as:
\begin{equation}
\begin{aligned}
\rm{\mathbf{H_{t}}} = \rm{\mathbf{ReLu}}(\rm{\mathbf{W}}\rm{\mathbf{H^{'}_{t}}} + \rm{\mathbf{U}}\rm{\mathbf{Y_{t}}}),
\end{aligned}
\end{equation}
where $\rm{\mathbf{W}}$ and $\rm{\mathbf{U}}$ are weight matrices, $t=1,2,\cdots,T$. $\rm{\mathbf{H^{'}}}$ will be padded to length $T$ when $T^{'}$ is smaller than $T$.
Outputs of the fusion layer are then fed to the top layer. 

\indent{}\textbf{Top Layer.} 
The top layer of the decoder is identical to the original Transformer decoder layer, which does not contain the positional attention layer compared to bottom layers. The outputs are fed to the softmax layer. 

Like other autoregressive models, FS-decoder has to generate translations through decoding algorithms such as greedy search and beam search. During decoding, bottom layers run in advance to prepare the inputs for the fusion layer, and then the fusion layer and top layer run autoregressively with the embedding of predicted token fed to the fusion layer.

\section{Related Work}
\citet{gu2017non} introduced the non-autoregressive Transformer model to accelerate the translation. \citet{lee2018deterministic} proposed a non-autoregressive sequence model based on iterative refinement, where the outputs of the decoder are fed back as inputs in the next iteration. \citet{guo2018non} proposed to enhance the decoder inputs with phrase-table lookup and embedding mapping. \citet{kaiser2018fast} used a sequence of autoregressively generated discrete latent variables as inputs of the decoder. Knowledge distillation \cite{hinton2015distilling,kim2016sequence} is a method for training a smaller and faster student network to perform better by learning from a teacher network, which is crucial in NAT models. \citet{gu2017non} applied Sequence-level knowledge distillation to eliminate the multimodality in the training corpus. \citet{li2018hint} further proposed to improve non-autoregressive models through distilling knowledge from intermediary hidden states and attention weights of autoregressive models. \\
\indent{}Apart from non-autoregressive translation, there are works toward speeding up the translation from other perspectives. \citet{wang2018semi} proposed the semi-autoregressive Transformer that generates a group of words in parallel at each time step. \citet{DBLP:journals/corr/abs-1810-13409} proposed the eager translation model that does not use the attention mechanism and has low latency. \citet{zhang2018accelerating} proposed the average attention network to accelerate decoding, which achieves significant speedup over the uncached Transformer. \citet{zhang2018speeding} proposed cube pruning to speedup the beam search for neural machine translation without damaging the translation quality.\\
\indent{}Sequence-level training techniques have been widely explored in autoregressive neural machine translation, where most works \cite{ranzato2015sequence,shen2016minimum,wu2016google,he2016dual,wu2017adversarial,yang2017improving} relied on reinforcement learning \cite{williams1992simple,sutton2000policy} to build the gradient estimator. Recently, techniques for sequence-level training with continuous objectives have been explored, including deterministic policy gradient algorithms \cite{gu2017trainable}, bag-of-words objective \cite{ma2018bag} and probabilistic n-gram matching \cite{shao2018greedy}. However, to the best of our knowledge, sequence-level training has not been applied to non-autoregressive models yet.\\
\indent{}The methods of variance reduction through focusing on the important parts of the distribution include importance sampling \cite{bengio2003quick,glynn1989importance} and complementary sum sampling \cite{botev2017complementary}. Importance sampling estimates the properties of a particular distribution through sampling on a different proposal distribution.
Complementary sum sampling reducdes the variance through suming over the important subset and estimating the rest via sampling.

{
\centering
\begin{table*}[t]
  \small
  \begin{center}
  \setlength{\tabcolsep}{1.5mm}{
    \begin{tabular}[b]{llrrrr|rrrr|rrrr}
    \toprule
    & & \multicolumn{4}{c|}{IWSLT'16 En-De} & \multicolumn{4}{c|}{WMT'16 En-Ro} & \multicolumn{4}{c}{WMT'14 En-De}  \\
    & & En$\rightarrow$ &  toks/s  & speedup  & secs/b & En$\rightarrow$ & Ro$\rightarrow$ & toks/s & speedup  & En$\rightarrow$ & De$\rightarrow$ & toks/s & speedup  \\
    \midrule
    \multirow{2}{*}{\rotatebox[origin=c]{90}{\scriptsize AR}}
        & b=1   & 28.13  & 45.3  &  1.09$\times$&  0.20& 31.53 & 31.35  & 45.6 &  1.23$\times$  & 23.67  & 28.04 & 33.7 &  1.13$\times$ \\
     & b=4   & 28.25   & 41.6&  1.00$\times$ &  0.20& 31.85 & 31.60  & 37.1 &  1.00$\times$ & 24.29  & 28.86 & 29.9&  1.00$\times$ \\  
    \midrule
    \multirow{2}{*}{\rotatebox[origin=c]{90}{\scriptsize NAT}}
        & FT & 26.52 & -- & 15.6 $\times$  &  --& 27.29 & 29.06 & -- & --& 17.69 & 21.47 & -- & --\\
        & FT+NPD & 28.16 & -- & 2.36 $\times$ &  --& 29.79 & 31.44 & -- & -- & 19.17 & 23.20  & -- & --\\  
    \midrule
     \multirow{2}{*}{\rotatebox[origin=c]{90}{\scriptsize IRNAT}}
    & iter=2       & 24.82 &  423.8  &  6.64 $\times$&  --& 27.10 & 28.15 &  332.7 &  7.68 $\times$  & 16.95 & 20.39 &  393.6&  8.77 $\times$  \\
        & adaptive       & 27.01  &  125.9  &  1.97 $\times$&  --& 29.66 & 30.30 &  118.3 &  2.73 $\times$  & 21.54 & 25.43 &  107.2 &  2.39 $\times$ \\
    \midrule
     \multirow{4}{*}{\rotatebox[origin=c]{90}{\scriptsize Our Models}}
     
     & NAT-base   & 24.13   & 350.2 &  8.42$\times$&  0.62& 25.96  & 26.49 & 349.0&  9.41$\times$  & 16.05 & 19.46  & 321.7&  10.76$\times$ \\
     & \  +REINFORCE   & 24.30    & 354.1&  8.51$\times$&  2.51  & 26.49& 27.20 & 346.7  &  9.35$\times$& 18.47 & 21.89 & 323.2 &  10.81$\times$ \\
     & \ +Reinforce-NAT   & 25.18   & 350.6 &  8.43$\times$&  13.40& 27.09  & 27.93 & 350.3 &  9.44$\times$ & 19.15 & 22.52  &320.9&  10.73$\times$\\
     \cmidrule{2-14}
      & FS-decoder(b=1)   & 27.58   & 168.7 &  4.06$\times$&  0.241  & 30.53 & 30.68  & 170.5 &  4.60$\times$ & 21.53  & 27.20 & 143.3&  4.79$\times$  \\
     & FS-decoder(b=4)   & 27.78   & 140.8 &  3.38$\times$&  0.241  & 30.57 & 30.83  & 137.1 &  3.70$\times$& 22.27  & 27.25 & 112.2&  3.75$\times$  \\

    \bottomrule
  \end{tabular}
  }
    \vspace{-2mm}
  \caption{Generation quality (4-gram BLEU), decoding efficiency (tokens/sec), speedup and training speed (seconds/batch). Decoding efficiency is measured sentence-by-sentence from the En$\rightarrow$ direction. Speedup is calculated over the autoregressive Transformer with beam size 4. NAT: non-autoregressive transformer models \cite{gu2017non}. IRNAT: iterative refinement for NAT \cite{lee2018deterministic}. AR: the autoregressive Transformer model. $b$: beam size. FS-decoder: fuse the sequential information into the top layer. NAT-base: our non-autoregressive baseline. +REINFORCE: finetune the NAT-base with REINFORCE according to Eq.(\ref{eq:nonauto_rf}). +Reinforce-NAT: finetune the NAT-base with Reinforce-NAT according to Eq.(\ref{eq:rfnat}).
  }
  \label{tab:bleu_performance2}
  \end{center}
    \vspace{-6mm}
\end{table*}    
}
\section{Experiments}
\subsection{Settings}
\noindent{}\textbf{Dataset.} We conduct experiments on three translation tasks\footnote{We release the source code in https://github.com/ictnlp/RSI-NAT}: IWSLT16 En$\rightarrow$De (196k pairs), WMT14 En$\leftrightarrow$De (4.5M pairs) and WMT16 En$\leftrightarrow$Ro (610k pairs). We use the preprocessed datasets released by \citet{lee2018deterministic}, where all sentences are tokenized and segmented into subword units using the BPE algorithm \cite{sennrich2015neural}. For all tasks, source and target languages share the vocabulary with size 40k. 
For WMT14 En-De, we employ newstest-2013 and newstest-2014 as development and test sets. 
For WMT16 En-Ro, we take newsdev-2016 and newstest-2016 as development and test sets. 
For IWSLT16 En-De, we use the test2013 for validation.

\vspace{5pt}
\noindent{}\textbf{Baselines.} We take the Transformer model \cite{vaswani2017attention} as the autoregressive baseline. The non-autoregressive model based on iterative refinement \cite{lee2018deterministic} is the non-autoregressive baseline, and we set the number of iterations to 2.

\vspace{5pt}
\noindent{}\textbf{Pre-train.} To evaluate the sequence-level training methods, we pre-train the NAT baseline first and then fine-tune the baseline model with GLEU \cite{wu2016google}, which outperforms other metrics in our experiments.
We stop the pre-train procedure, when training steps are more than 300k and no further improvements on the validation set are observed in last 100k steps.

\vspace{5pt}
\noindent{}\textbf{Hyperparameters.} We closely follow the setting of \citet{gu2017non} and \citet{lee2018deterministic}. In IWSLT16 En-De, we use the small model ($d_{\rm model}$=$278$, $d_{\rm hidden}$=$507$, $n_{\rm layer}$=$5$, $n_{\rm head}$=$2$, $p_{\rm dropout}$=$0.1$, $t_{\rm warmup}$=$746$). For experiments on WMT datasets, we use the base Transformer \citet{vaswani2017attention} ($d_{\rm model}$=$512$, $d_{\rm hidden}$=$512$, $n_{\rm layer}$=$6$, $n_{\rm head}$=$8$, $p_{\rm dropout}$=$0.1$, $\rm t_{warmup}$=$16000$). 
The traversing count $k$ and the sampling times $n$ in algorithm \ref{alg:1} are respectively set to 5 and 20. 
We use Adam \cite{DBLP:journals/corr/KingmaB14} for the optimization. 
During decoding, we remove any token that is generated repeatly. The decoding speed is measured on a single Geforce GTX TITAN X.

\vspace{5pt}
\noindent{}\textbf{Knowledge Distillation.} Knowledge distillation \cite{kim2016sequence,hinton2015distilling} is proved to be crucial for successfully training NAT models \cite{gu2017non,li2018hint}. 
For all the translation tasks, we apply sequence-level knowledge distillation to construct the distillation corpus where the target side of the training corpus is replaced by the output of an autoregressive Transformer model. We use original corpora to train the autoregressive baseline and distillation corpora to train other models. 
\subsection{Main Results}
We compare our models with the NAT \cite{gu2017non} and the IRNAT \cite{lee2018deterministic}. 
Table \ref{tab:bleu_performance2} shows the experiment results. We observe that models based on sequence-level training approaches, including REINFORCE and Reinforce-NAT, significantly surpass the NAT baseline on BLEU without damaging the decoding speed. The Reinforce-NAT model outperforms the REINFORCE model in terms of BLEU points. On WMT14 En$\leftrightarrow$De, the Reinforce-NAT model achieves significant improvements by more than 3 BLEU points and outperforms NAT(FT) \cite{gu2017non} and IRNAT(iteration=2) \cite{lee2018deterministic}. The above results demonstrate the effectiveness of sequence-level training and prove the strong ability of Reinforce-NAT. The experiment on the FS-decoder show that it brings huge BLEU improvements over the NAT baseline and even achieves comparable performance to the autoregressive Transformer with considerable speedup, which proves the capacity of the FS-decoder.

\subsection{Training Speed}
Table \ref{tab:bleu_performance2} shows the training time per batch of our methods. Sequence-level training methods (i.e., REINFORCE and Reinforce-NAT) are slower than the word-level training. The bottleneck lies in the calculation of the reward (i.e., GLEU), which takes place in CPU and can be accelerated by multi-processing. Besides, these methods are only utilized to fine-tune the baseline model and take
less than 10,000 batches to converge, which make the relatively low training speed affordable.

\subsection{Effect of top-$k$ size in Reinforce-NAT}
\begin{figure}[h]
  \begin{center}
    \includegraphics[scale=0.5]{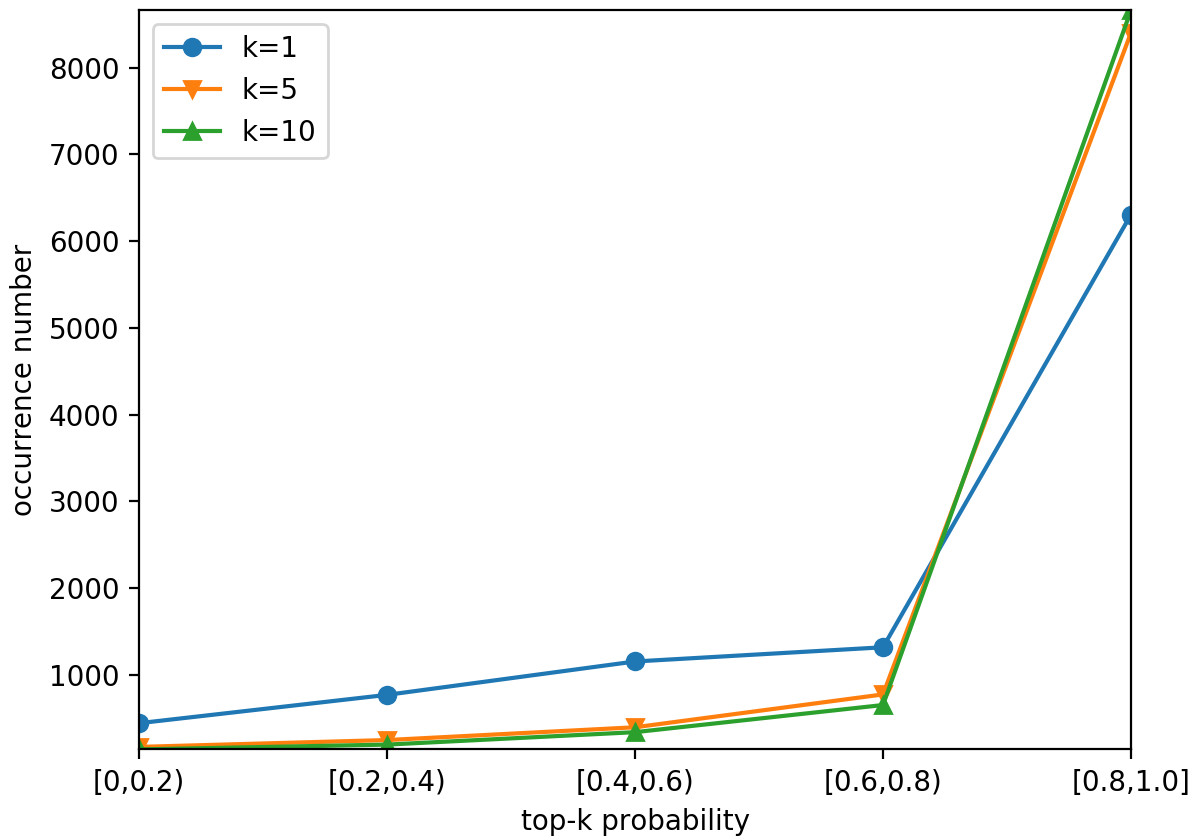}
    \caption{top-$k$ probability distributions for $k$=$1$, $5$ and $10$}
    \label{fig:3}
  \end{center}
  \vspace{-0.5em}
\end{figure}

The Reinforce-NAT is proposed on the basis that the top-$k$ words can occupy the central part of the probability distribution. However, it remains unknown which $k$ is appropriate for us. A large $k$ will slow down the training, and a small $k$ will be not enough to dominate the probability distribution. We statistically and experimentally analyze the choice of $k$ in Reinforce-NAT. 
\indent{}We respectively set k to $1$, $5$ and $10$ and record the top-$k$ probabilities in 10,000 target word predictions. Figure \ref{fig:3} and Table \ref{tab1} illustrate the statistical properties of top-$k$ probabilities. In figure \ref{fig:3}, the x-axis divides the probability distribution into 5 intervals, and the y-axis indicates the number of times that the top-$k$ probabilities are within this interval. In Table \ref{tab1}, we estimate the expection of top-$k$ probabilities for different $k$. We find that $k=5$ is a desirable choice that can cover a large portion of the probability distribution, and the marginal utility for a larger $k$ is limitted.

\begin{table}[!htbp]
\centering
\begin{tabular}{c|c|c|c|c|c}
\toprule
$k$& 1& 5& 10& 100& 1000\\
\hline
E[$P_k$]&0.818&0.916&0.929&0.948&0.968\\
\bottomrule
\end{tabular}
\caption{top-$k$ probability expection for $k$=$1$, $5$, $10$, $100$, $1000$}
\label{tab1}
\end{table}

\begin{figure}[h]
  \begin{center}
    \includegraphics[scale=0.5]{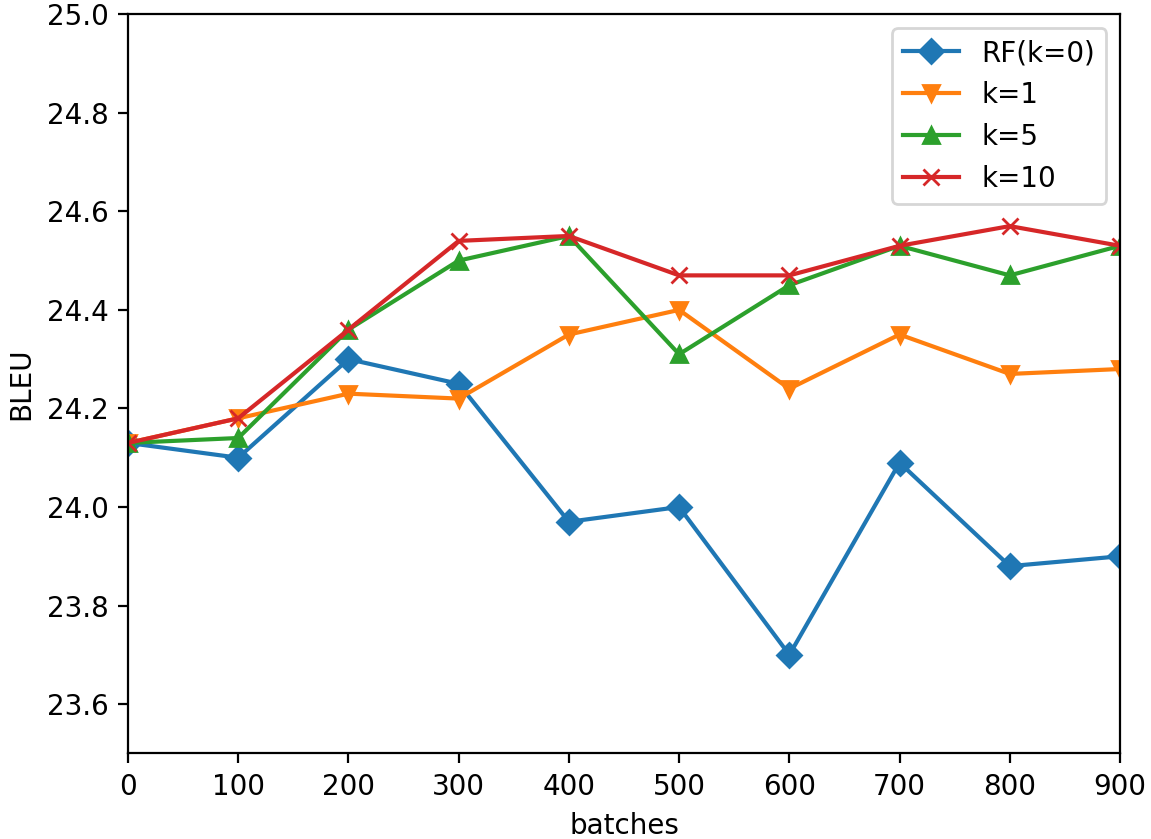}
    \caption{training curves for $k=0$, $1$, $5$ and $10$.}
    \label{fig:4}
  \end{center}
  \vspace{-0.5em}
\end{figure}

\indent{}We further conduct experiments on IWSLT16 En$\rightarrow$De to confirm the conclusion. We respectively set k to $0$, $1$, $5$ and $10$ in Reinforce-NAT and draw training curves. Figure \ref{fig:4} shows that REINFORCE($k=0$) is very unstable in the training, and greater $k$ in Reinforce-NAT generally leads to better performance. In line with our previous conclusion, $k=5$ is an ideal choice since it does not have a large performance gap between larger $k$. 

\subsection{Performance over Different Lengths}
Table \ref{tab:bleu_performance2} shows that the performance of Reinforce-NAT varies with datasets. Though IWSLT16 En$\rightarrow$De and WMT14 En$\rightarrow$De have the same language pair, Reinforce-NAT achieves an improvement of more than 3 BLEU points on WMT14 but only have about 1.0 BLEU points improvement on IWSLT16. We attribute this phenomenon to the length difference between two datasets. The WMT14 En$\rightarrow$De dataset is in the news-domain, whose sentences are statistically longer than the spoken-domain IWSLT16 En$\rightarrow$De dataset.

\begin{figure}[htb]
  \begin{center}
    \includegraphics[scale=0.5]{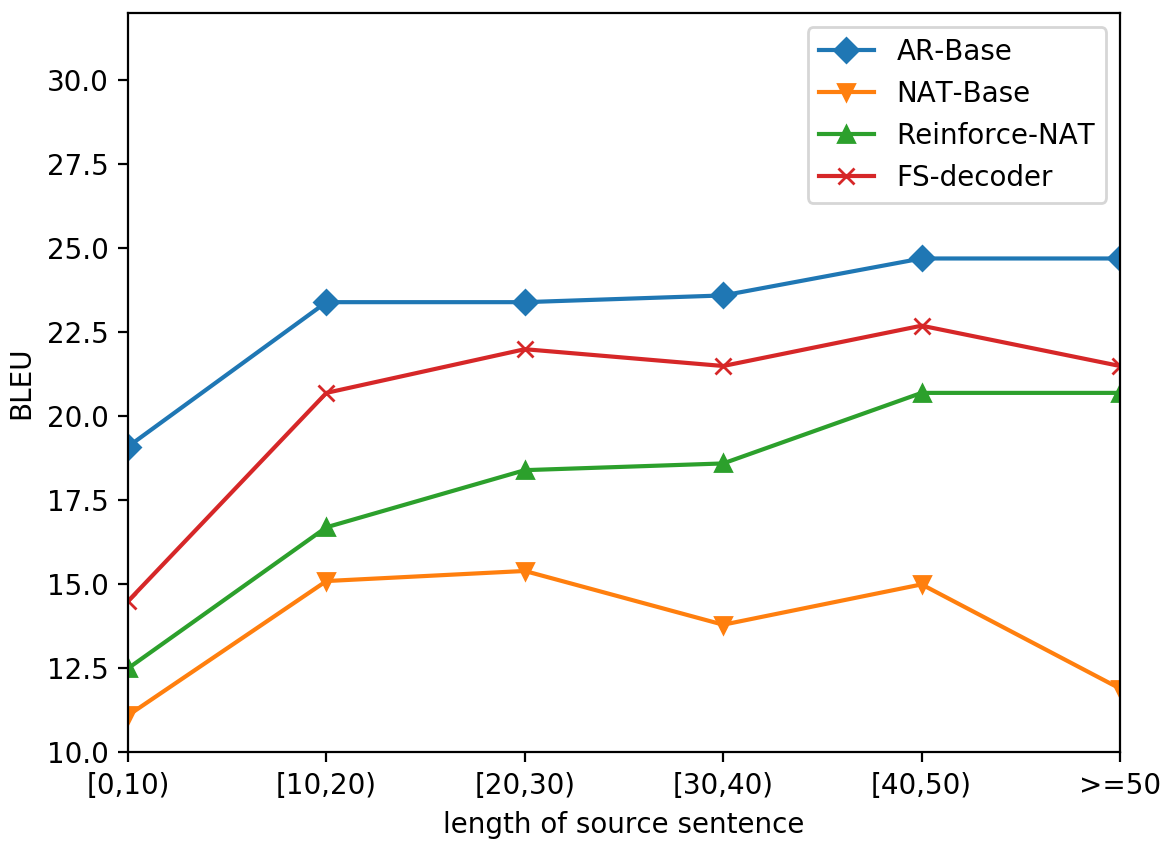}
    \caption{The BLEU scores on the validation set of WMT14 En$\rightarrow$De over sentences in different length buckets. The beam size of FS-decoder and AR-Base is 1.}
    \label{fig:length}
  \end{center}
  \vspace{-0.5em}
\end{figure}

Figure \ref{fig:length} shows BLEU scores over sentences in different length buckets. The BLEU scores of NAT-Base have a distinct decrease when the sentence length is over 40, while other models perform well on long sentences. It confirms that NAT models are weak in translating long sentences and our solutions can effectively improve the performance of NAT models on long sentences through leveraging sequential information.

\subsection{Case Study}
\begin{table*}[t]
\small
\centering
\begin{tabular}{r|l}
\toprule
\multirow{2}{*}{Source} & und noch tragischer ist , dass es Oxford war - eine Universität , die nicht nur 14 Tory-Premierminister \\&hervorbrachte , sondern sich bis heute hinter einem unverdienten Ruf von Gleichberechtigung und \\&Gedankenfreiheit versteckt .\\
\hline
\multirow{2}{*}{Target} & even more tragic is that it was Oxford , which not only produced 14 Tory prime ministers , \\& but , to this day , hides behind an ill-deserved reputation for equality and freedom of thought . \\
\hline
\multirow{2}{*}{NAT-Base} & and more more more more that it was Oxford - a university that not not only only TTory Prime Minister ,\\& but has has to hidden hidden behind an unfounded reputation of equality and freedom of thought .\\
\hline
\multirow{2}{*}{Reinforce-NAT} & and more more tragic is that it was Oxford - a university that did not only produce 14 Tory Prime Minister \\&, but has still to be hidden behind an unfied reputation of equality and freedom of thought . \\
\hline
\multirow{2}{*}{FS-decoder} & and even more tragic , it was Oxford - a university that produced not only 14 Tory Prime Minister , \\&but still hidden behind an unbridled reputation of equality and freedom of thought .\\
\hline
\multirow{2}{*}{AR-Base} & and , more tragic , Oxford was - a university that not only produced 14 Tory Prime Minister , \\&but still hidden behind an unprecedented reputation for equality and freedom of thought .
\\ 
\bottomrule
\end{tabular}
\caption{A translation case on WMT14 De$\rightarrow$En task. Over-translation and under-translation errors occur in the translation of NAT-Base. }
\label{tab:case_study}
\end{table*}
In Table \ref{tab:case_study}, we present a translation case from the validation set of WMT14 De$\rightarrow$En. The case shows that the translation quality rise in the order of NAT-Base, +Reinforce-NAT, FS-decoder to AR-Base and the performance gap is large between NAT-Base and other models.
Particularly, NAT models suffer from over-translation and under-translation when translating long sentences, which is efficiently alleviated by Reinforce-NAT and RF-Decoder.

\section{Conclusion}
In this paper, we aim to retrieve the sequential information for NAT models to enhance their translation ability while preserving fast-decoding property.
Firstly, we propose a sequence-level training method based on a novel reinforcement algorithm for NAT (Reinforce-NAT), which significantly improves the performance of NAT models without decelerating the decoding speed. Secondly, we propose an innovative Transformer decoder named FS-decoder to fuse the target sequential information into the top layer of the decoder, which achieves comparable performance to the Transformer and still maintains substantial speedup. 

In the future, we plan to investigate better methods to leverage the sequential information. We believe that the following two directions are worth study. First, exploiting other sequence-level training objectives like bag-of-words \cite{ma2018bag}. Second, using sequential information distilled from the autoregressive teacher model to guide the training of the student non-autoregressive model.
\section{Acknowledgments}
We thank the anonymous reviewers for their insightful comments. This work was supported by National Natural Science Foundation of China (NO.61662077, NO.61876174) and National Key R\&D Program of China (NO.YS2017YFGH001428).

\bibliography{acl2019}

\begin{thebibliography}{41}
\expandafter\ifx\csname natexlab\endcsname\relax\def\natexlab#1{#1}\fi

\bibitem[{Bahdanau et~al.(2016)Bahdanau, Brakel, Xu, Goyal, Lowe, Pineau,
  Courville, and Bengio}]{bahdanau2016actor}
Dzmitry Bahdanau, Philemon Brakel, Kelvin Xu, Anirudh Goyal, Ryan Lowe, Joelle
  Pineau, Aaron Courville, and Yoshua Bengio. 2016.
\newblock An actor-critic algorithm for sequence prediction.
\newblock \emph{arXiv preprint arXiv:1607.07086}.

\bibitem[{Bahdanau et~al.(2014)Bahdanau, Cho, and Bengio}]{bahdanau2014neural}
Dzmitry Bahdanau, Kyunghyun Cho, and Yoshua Bengio. 2014.
\newblock Neural machine translation by jointly learning to align and
  translate.
\newblock \emph{arXiv preprint arXiv:1409.0473}.

\bibitem[{Bengio et~al.(2003)Bengio, Sen{\'e}cal et~al.}]{bengio2003quick}
Yoshua Bengio, Jean-S{\'e}bastien Sen{\'e}cal, et~al. 2003.
\newblock Quick training of probabilistic neural nets by importance sampling.
\newblock In \emph{AISTATS}, pages 1--9.

\bibitem[{Botev et~al.(2017)Botev, Zheng, and Barber}]{botev2017complementary}
Aleksandar Botev, Bowen Zheng, and David Barber. 2017.
\newblock Complementary sum sampling for likelihood approximation in large
  scale classification.
\newblock In \emph{Artificial Intelligence and Statistics}, pages 1030--1038.

\bibitem[{Cho et~al.(2014)Cho, Van~Merri{\"e}nboer, Gulcehre, Bahdanau,
  Bougares, Schwenk, and Bengio}]{cho2014learning}
Kyunghyun Cho, Bart Van~Merri{\"e}nboer, Caglar Gulcehre, Dzmitry Bahdanau,
  Fethi Bougares, Holger Schwenk, and Yoshua Bengio. 2014.
\newblock Learning phrase representations using rnn encoder-decoder for
  statistical machine translation.
\newblock \emph{arXiv preprint arXiv:1406.1078}.

\bibitem[{Glynn and Iglehart(1989)}]{glynn1989importance}
Peter~W Glynn and Donald~L Iglehart. 1989.
\newblock Importance sampling for stochastic simulations.
\newblock \emph{Management Science}, 35(11):1367--1392.

\bibitem[{Gu et~al.(2017{\natexlab{a}})Gu, Bradbury, Xiong, Li, and
  Socher}]{gu2017non}
Jiatao Gu, James Bradbury, Caiming Xiong, Victor~OK Li, and Richard Socher.
  2017{\natexlab{a}}.
\newblock Non-autoregressive neural machine translation.
\newblock \emph{arXiv preprint arXiv:1711.02281}.

\bibitem[{Gu et~al.(2017{\natexlab{b}})Gu, Cho, and Li}]{gu2017trainable}
Jiatao Gu, Kyunghyun Cho, and Victor~OK Li. 2017{\natexlab{b}}.
\newblock Trainable greedy decoding for neural machine translation.
\newblock In \emph{Proceedings of the 2017 Conference on Empirical Methods in
  Natural Language Processing}, pages 1968--1978.

\bibitem[{Guo et~al.(2018)Guo, Tan, He, Qin, Xu, and Liu}]{guo2018non}
Junliang Guo, Xu~Tan, Di~He, Tao Qin, Linli Xu, and Tie-Yan Liu. 2018.
\newblock Non-autoregressive neural machine translation with enhanced decoder
  input.
\newblock \emph{arXiv preprint arXiv:1812.09664}.

\bibitem[{He et~al.(2016)He, Xia, Qin, Wang, Yu, Liu, and Ma}]{he2016dual}
Di~He, Yingce Xia, Tao Qin, Liwei Wang, Nenghai Yu, Tieyan Liu, and Wei-Ying
  Ma. 2016.
\newblock Dual learning for machine translation.
\newblock In \emph{Advances in Neural Information Processing Systems}, pages
  820--828.

\bibitem[{Hinton et~al.(2015)Hinton, Vinyals, and Dean}]{hinton2015distilling}
Geoffrey Hinton, Oriol Vinyals, and Jeff Dean. 2015.
\newblock Distilling the knowledge in a neural network.
\newblock \emph{arXiv preprint arXiv:1503.02531}.

\bibitem[{Kaiser et~al.(2018)Kaiser, Roy, Vaswani, Pamar, Bengio, Uszkoreit,
  and Shazeer}]{kaiser2018fast}
{\L}ukasz Kaiser, Aurko Roy, Ashish Vaswani, Niki Pamar, Samy Bengio, Jakob
  Uszkoreit, and Noam Shazeer. 2018.
\newblock Fast decoding in sequence models using discrete latent variables.
\newblock \emph{arXiv preprint arXiv:1803.03382}.

\bibitem[{Kim and Rush(2016)}]{kim2016sequence}
Yoon Kim and Alexander~M Rush. 2016.
\newblock Sequence-level knowledge distillation.
\newblock In \emph{Proceedings of the 2016 Conference on Empirical Methods in
  Natural Language Processing}, pages 1317--1327.

\bibitem[{Kingma and Ba(2014)}]{DBLP:journals/corr/KingmaB14}
Diederik~P. Kingma and Jimmy Ba. 2014.
\newblock \href {http://arxiv.org/abs/1412.6980} {Adam: {A} method for
  stochastic optimization}.
\newblock \emph{CoRR}, abs/1412.6980.

\bibitem[{Lee et~al.(2018)Lee, Mansimov, and Cho}]{lee2018deterministic}
Jason Lee, Elman Mansimov, and Kyunghyun Cho. 2018.
\newblock Deterministic non-autoregressive neural sequence modeling by
  iterative refinement.
\newblock \emph{arXiv preprint arXiv:1802.06901}.

\bibitem[{Li et~al.(2018)Li, He, Tian, Qin, Wang, and Liu}]{li2018hint}
Zhuohan Li, Di~He, Fei Tian, Tao Qin, Liwei Wang, and Tie-Yan Liu. 2018.
\newblock Hint-based training for non-autoregressive translation.

\bibitem[{Ma et~al.(2018)Ma, Sun, Wang, and Lin}]{ma2018bag}
Shuming Ma, Xu~Sun, Yizhong Wang, and Junyang Lin. 2018.
\newblock Bag-of-words as target for neural machine translation.
\newblock \emph{arXiv preprint arXiv:1805.04871}.

\bibitem[{Ng et~al.(1999)Ng, Harada, and Russell}]{Ng1999PolicyIU}
Andrew~Y. Ng, Daishi Harada, and Stuart~J. Russell. 1999.
\newblock Policy invariance under reward transformations: Theory and
  application to reward shaping.
\newblock In \emph{ICML}.

\bibitem[{Papineni et~al.(2002)Papineni, Roukos, Ward, and
  Zhu}]{papineni2002bleu}
Kishore Papineni, Salim Roukos, Todd Ward, and Wei-Jing Zhu. 2002.
\newblock Bleu: a method for automatic evaluation of machine translation.
\newblock In \emph{Proceedings of the 40th annual meeting on association for
  computational linguistics}, pages 311--318. Association for Computational
  Linguistics.

\bibitem[{Press and Smith(2018)}]{DBLP:journals/corr/abs-1810-13409}
Ofir Press and Noah~A. Smith. 2018.
\newblock \href {http://arxiv.org/abs/1810.13409} {You may not need attention}.
\newblock \emph{CoRR}, abs/1810.13409.

\bibitem[{Ranzato et~al.(2015)Ranzato, Chopra, Auli, and
  Zaremba}]{ranzato2015sequence}
Marc'Aurelio Ranzato, Sumit Chopra, Michael Auli, and Wojciech Zaremba. 2015.
\newblock Sequence level training with recurrent neural networks.
\newblock \emph{arXiv preprint arXiv:1511.06732}.

\bibitem[{Sennrich et~al.(2016)Sennrich, Haddow, and
  Birch}]{sennrich2015neural}
Rico Sennrich, Barry Haddow, and Alexandra Birch. 2016.
\newblock \href {http://www.aclweb.org/anthology/P16-1162} {Neural machine
  translation of rare words with subword units}.
\newblock In \emph{Proceedings of the 54th Annual Meeting of the Association
  for Computational Linguistics (Volume 1: Long Papers)}, pages 1715--1725,
  Berlin, Germany. Association for Computational Linguistics.

\bibitem[{Shao et~al.(2018)Shao, Chen, and Feng}]{shao2018greedy}
Chenze Shao, Xilin Chen, and Yang Feng. 2018.
\newblock Greedy search with probabilistic n-gram matching for neural machine
  translation.
\newblock In \emph{Proceedings of the 2018 Conference on Empirical Methods in
  Natural Language Processing}, pages 4778--4784.

\bibitem[{Shen et~al.(2016)Shen, Cheng, He, He, Wu, Sun, and
  Liu}]{shen2016minimum}
Shiqi Shen, Yong Cheng, Zhongjun He, Wei He, Hua Wu, Maosong Sun, and Yang Liu.
  2016.
\newblock Minimum risk training for neural machine translation.
\newblock In \emph{Proceedings of the 54th Annual Meeting of the Association
  for Computational Linguistics (Volume 1: Long Papers)}, volume~1, pages
  1683--1692.

\bibitem[{Snover et~al.(2006)Snover, Dorr, Schwartz, Micciulla, and
  Makhoul}]{snover2006study}
Matthew Snover, Bonnie Dorr, Richard Schwartz, Linnea Micciulla, and John
  Makhoul. 2006.
\newblock A study of translation edit rate with targeted human annotation.
\newblock In \emph{Proceedings of association for machine translation in the
  Americas}, volume 200.

\bibitem[{Sutskever et~al.(2014)Sutskever, Vinyals, and
  Le}]{sutskever2014sequence}
Ilya Sutskever, Oriol Vinyals, and Quoc~V Le. 2014.
\newblock Sequence to sequence learning with neural networks.
\newblock In \emph{Advances in neural information processing systems}, pages
  3104--3112.

\bibitem[{Sutton et~al.(2000)Sutton, McAllester, Singh, and
  Mansour}]{sutton2000policy}
Richard~S Sutton, David~A McAllester, Satinder~P Singh, and Yishay Mansour.
  2000.
\newblock Policy gradient methods for reinforcement learning with function
  approximation.
\newblock In \emph{Advances in neural information processing systems}, pages
  1057--1063.

\bibitem[{Sutton(1984)}]{sutton1984temporal}
Richard~Stuart Sutton. 1984.
\newblock Temporal credit assignment in reinforcement learning.

\bibitem[{Vaswani et~al.(2017)Vaswani, Shazeer, Parmar, Uszkoreit, Jones,
  Gomez, Kaiser, and Polosukhin}]{vaswani2017attention}
Ashish Vaswani, Noam Shazeer, Niki Parmar, Jakob Uszkoreit, Llion Jones,
  Aidan~N Gomez, {\L}ukasz Kaiser, and Illia Polosukhin. 2017.
\newblock Attention is all you need.
\newblock In \emph{Advances in Neural Information Processing Systems}, pages
  6000--6010.

\bibitem[{Wang et~al.(2018)Wang, Zhang, and Chen}]{wang2018semi}
Chunqi Wang, Ji~Zhang, and Haiqing Chen. 2018.
\newblock Semi-autoregressive neural machine translation.
\newblock \emph{arXiv preprint arXiv:1808.08583}.

\bibitem[{Wang et~al.(2019)Wang, Tian, He, Qin, Zhai, and Liu}]{wang2019non}
Yiren Wang, Fei Tian, Di~He, Tao Qin, ChengXiang Zhai, and Tie-Yan Liu. 2019.
\newblock Non-autoregressive machine translation with auxiliary regularization.
\newblock \emph{arXiv preprint arXiv:1902.10245}.

\bibitem[{Weaver and Tao(2013)}]{article}
Lex Weaver and Nigel Tao. 2013.
\newblock The optimal reward baseline for gradient-based reinforcement
  learning.
\newblock \emph{Processings of the Seventeeth Conference on Uncertainty in
  Artificial Intelligence}.

\bibitem[{Williams(1992)}]{williams1992simple}
Ronald~J Williams. 1992.
\newblock Simple statistical gradient-following algorithms for connectionist
  reinforcement learning.
\newblock In \emph{Reinforcement Learning}, pages 5--32. Springer.

\bibitem[{Williams and Zipser(1989)}]{williams1989learning}
Ronald~J Williams and David Zipser. 1989.
\newblock A learning algorithm for continually running fully recurrent neural
  networks.
\newblock \emph{Neural computation}, 1(2):270--280.

\bibitem[{Wu et~al.(2018)Wu, Tian, Qin, Lai, and Liu}]{wu2018study}
Lijun Wu, Fei Tian, Tao Qin, Jianhuang Lai, and Tie-Yan Liu. 2018.
\newblock A study of reinforcement learning for neural machine translation.
\newblock \emph{arXiv preprint arXiv:1808.08866}.

\bibitem[{Wu et~al.(2017)Wu, Xia, Zhao, Tian, Qin, Lai, and
  Liu}]{wu2017adversarial}
Lijun Wu, Yingce Xia, Li~Zhao, Fei Tian, Tao Qin, Jianhuang Lai, and Tie-Yan
  Liu. 2017.
\newblock Adversarial neural machine translation.
\newblock \emph{arXiv preprint arXiv:1704.06933}.

\bibitem[{Wu et~al.(2016)Wu, Schuster, Chen, Le, Norouzi, Macherey, Krikun,
  Cao, Gao, Macherey et~al.}]{wu2016google}
Yonghui Wu, Mike Schuster, Zhifeng Chen, Quoc~V Le, Mohammad Norouzi, Wolfgang
  Macherey, Maxim Krikun, Yuan Cao, Qin Gao, Klaus Macherey, et~al. 2016.
\newblock Google's neural machine translation system: Bridging the gap between
  human and machine translation.
\newblock \emph{arXiv preprint arXiv:1609.08144}.

\bibitem[{Yang et~al.(2017)Yang, Chen, Wang, and Xu}]{yang2017improving}
Zhen Yang, Wei Chen, Feng Wang, and Bo~Xu. 2017.
\newblock Improving neural machine translation with conditional sequence
  generative adversarial nets.
\newblock \emph{arXiv preprint arXiv:1703.04887}.

\bibitem[{Yu et~al.(2017)Yu, Zhang, Wang, and Yu}]{yu2017seqgan}
Lantao Yu, Weinan Zhang, Jun Wang, and Yong Yu. 2017.
\newblock Seqgan: Sequence generative adversarial nets with policy gradient.
\newblock In \emph{AAAI}, pages 2852--2858.

\bibitem[{Zhang et~al.(2018{\natexlab{a}})Zhang, Xiong, and
  Su}]{zhang2018accelerating}
Biao Zhang, Deyi Xiong, and Jinsong Su. 2018{\natexlab{a}}.
\newblock Accelerating neural transformer via an average attention network.
\newblock \emph{arXiv preprint arXiv:1805.00631}.

\bibitem[{Zhang et~al.(2018{\natexlab{b}})Zhang, Huang, Feng, Shen, and
  Liu}]{zhang2018speeding}
Wen Zhang, Liang Huang, Yang Feng, Lei Shen, and Qun Liu. 2018{\natexlab{b}}.
\newblock Speeding up neural machine translation decoding by cube pruning.
\newblock In \emph{Proceedings of the 2018 Conference on Empirical Methods in
  Natural Language Processing}, pages 4284--4294.

\end{thebibliography}
\bibliographystyle{acl_natbib}
\newpage
\ 
\newpage
\appendix
\section{Supplemental Material}
Proof for Eq.(\ref{eq:nonauto_reduce}):
\begin{equation}
\begin{aligned}
-\nabla_{\theta}\mathrm{L}_\theta=&\sum_{\bm{\mathrm{Y}}}\nabla_{\theta}\prod_{t=1}^{T}p(y_t|\bm{X},\theta) \cdot r(\bm{\mathrm{Y}})\\
=&\sum_{\bm{\mathrm{Y}}}\sum_{t=1}^{T}\nabla_{\theta}p(y_t|\bm{X},\theta)\cdot\prod_{i=1}^{t-1}p(y_i|\bm{X},\theta)\cdot\prod_{j=t+1}^{T}p(y_j|\bm{X},\theta)\cdot r(\bm{\mathrm{Y}})\\
=&\sum_{t=1}^{T}\sum_{\bm{\mathrm{Y}}}\nabla_{\theta}p(y_t|\bm{X},\theta)\cdot\prod_{i=1}^{t-1}p(y_i|\bm{X},\theta)\cdot\prod_{j=t+1}^{T}p(y_j|\bm{X},\theta)\cdot r(\bm{\mathrm{Y}})\\
=&\sum_{t=1}^{T}\sum_{{y_t}}\nabla_{\theta}p(y_t|\bm{X},\theta)\cdot\sum_{\bm{\mathrm{y_{1:t-1}}}}\sum_{\bm{\mathrm{y_{t+1:T}}}}\prod_{i=1}^{t-1}p(y_i|\bm{X},\theta)\cdot\prod_{j=t+1}^{T}p(y_j|\bm{X},\theta)\cdot r(\bm{\mathrm{Y}})\\
=&\sum_{t=1}^{T}\sum_{{y_t}}\nabla_{\theta}p(y_t|\bm{X},\theta)\cdot\mathop{\mathbb{E}}\limits_{y_{1:t-1}}\mathop{\mathbb{E}}\limits_{y_{t+1:T}}r(\bm{\mathrm{Y}}).\\
=&\sum_{t=1}^{T}\sum_{y_t} \nabla_{\theta} p(y_{t}|\bm{\mathrm{X}},\theta) \cdot r(y_t)
\end{aligned}
\end{equation}
\end{document}